\documentclass[conference]{IEEEtran}
\IEEEoverridecommandlockouts
\usepackage{cite}

\usepackage{amsmath,amssymb,amsfonts,mathtools,bm}
\usepackage{amsthm}
\usepackage{algorithm}
\usepackage{algorithmic}
\usepackage{graphicx}
\usepackage{textcomp}
\usepackage{xcolor,float}
\usepackage{tabularx}
\usepackage{textcomp}
\usepackage{diagbox}
\usepackage{svg}
\usepackage{booktabs}
\usepackage{subfigure}
\usepackage{hyperref}

\allowdisplaybreaks
\renewcommand{\baselinestretch}{1}

\begin{document}

\title{Trapezoidal Gradient Descent for Effective Reinforcement Learning in Spiking Networks}
\author{
\IEEEauthorblockN{
Yuhao Pan\IEEEauthorrefmark{1},
Xiucheng Wang\IEEEauthorrefmark{2},
Nan Cheng\IEEEauthorrefmark{2},
Qi Qiu\IEEEauthorrefmark{1},\\
}
\IEEEauthorblockA{
\IEEEauthorrefmark{1}School of Electronic Engineering, Xidian University, Xi'an, 710071, China\\
\IEEEauthorrefmark{2}School of Telecommunications Engineering, Xidian University, Xi'an, 710071, China \\
Email: \{yhpan, xcwang\_1, qiuqi\}@stu.xidian.edu.cn, dr.nan.cheng@ieee.org}}

    \maketitle

\IEEEdisplaynontitleabstractindextext

\IEEEpeerreviewmaketitle

\begin{abstract}

With the rapid development of artificial intelligence technology, the field of reinforcement learning has continuously achieved breakthroughs in both theory and practice. However, traditional reinforcement learning algorithms often entail high energy consumption during interactions with the environment. Spiking Neural Network (SNN), with their low energy consumption characteristics and performance comparable to deep neural networks, have garnered widespread attention. To reduce the energy consumption of practical applications of reinforcement learning, researchers have successively proposed the Pop-SAN and MDC-SAN algorithms. Nonetheless, these algorithms use rectangular functions to approximate the spike network during the training process, resulting in low sensitivity, thus indicating room for improvement in the training effectiveness of SNN. Based on this, we propose a trapezoidal approximation gradient method to replace the spike network, which not only preserves the original stable learning state but also enhances the model’s adaptability and response sensitivity under various signal dynamics. Simulation results show that the improved algorithm, using the trapezoidal approximation gradient to replace the spike network, achieves better convergence speed and performance compared to the original algorithm and demonstrates good training stability.


\end{abstract}

\begin{IEEEkeywords}
SNN, reinforcement learning, spike network, trapezoidal function

\end{IEEEkeywords}

\section{Introduction}


Reinforcement learning (RL), as a significant branch of artificial intelligence, has demonstrated extensive application prospects in fields such as autonomous driving and robotics control \cite{arulkumaran2017deep, cheng2024enhanced, wang2022digital,wang2022joint,cheng2023ai}. RL focuses on solving sequential decision-making problems faced by agents interacting with their environments, aiming to maximize expected cumulative rewards through strategic policy learning. Despite the strong performance of deep reinforcement learning (DRL) in various control tasks \cite{zhou2020deep}, their practical implementation often demands substantial computational resources, limiting their sustainability in long-term applications.

Fortunately, the rise of Spiking Neural Networks (SNNs) offer a novel solution to the energy consumption challenges posed by DRL. SNNs significantly reduce energy consumption and extend the runtime of control tasks through their unique pulse coding and event-driven mechanisms. SNNs aim to bridge neuroscience and machine learning by utilizing computational models that closely mimic biological neurons \cite{tavanaei2019deep}. Their capabilities in processing complex temporal data, operating at low power, and their deep physiological underpinnings have garnered widespread academic attention, positioning them as the next generation of neural networks.

To address the high energy consumption of DRL, researchers have begun exploring the potential of integrating low-energy SNNs with RL to enhance the endurance and efficiency of agents. Patel et al. pioneered the application of conversion methods from ANN to SNN in training Deep Q Networks using RL \cite{patel2019improved}. However, this conversion process requires adjusting network weights, which can lead to decreased performance of the converted algorithm. Moreover, prolonged simulations increase energy consumption and inference latency. Subsequently, Tang et al. introduced the Population-coded spiking actor network (Pop-SAN) based on the Actor-Critic (AC) framework \cite{tang2021deep}, which consists of a spiking actor network and a deep critic network. This algorithm uses rectangular functions to approximate the non-differentiable parts of the spiking network, effectively reducing energy consumption in continuous control tasks. Zhang et al. improved the LIF neurons in Pop-SAN, introduced more complex Dynamic Neurons, and proposed the multiscale dynamic coding improved spiking actor network (MDC-SAN) \cite{zhang2022multi}. This method increases precision in continuous control tasks and employs rectangular functions, like Tang's approach, to approximate gradients during training.



\begin{figure*}
  \centering
  \includegraphics[width=1.55\columnwidth]{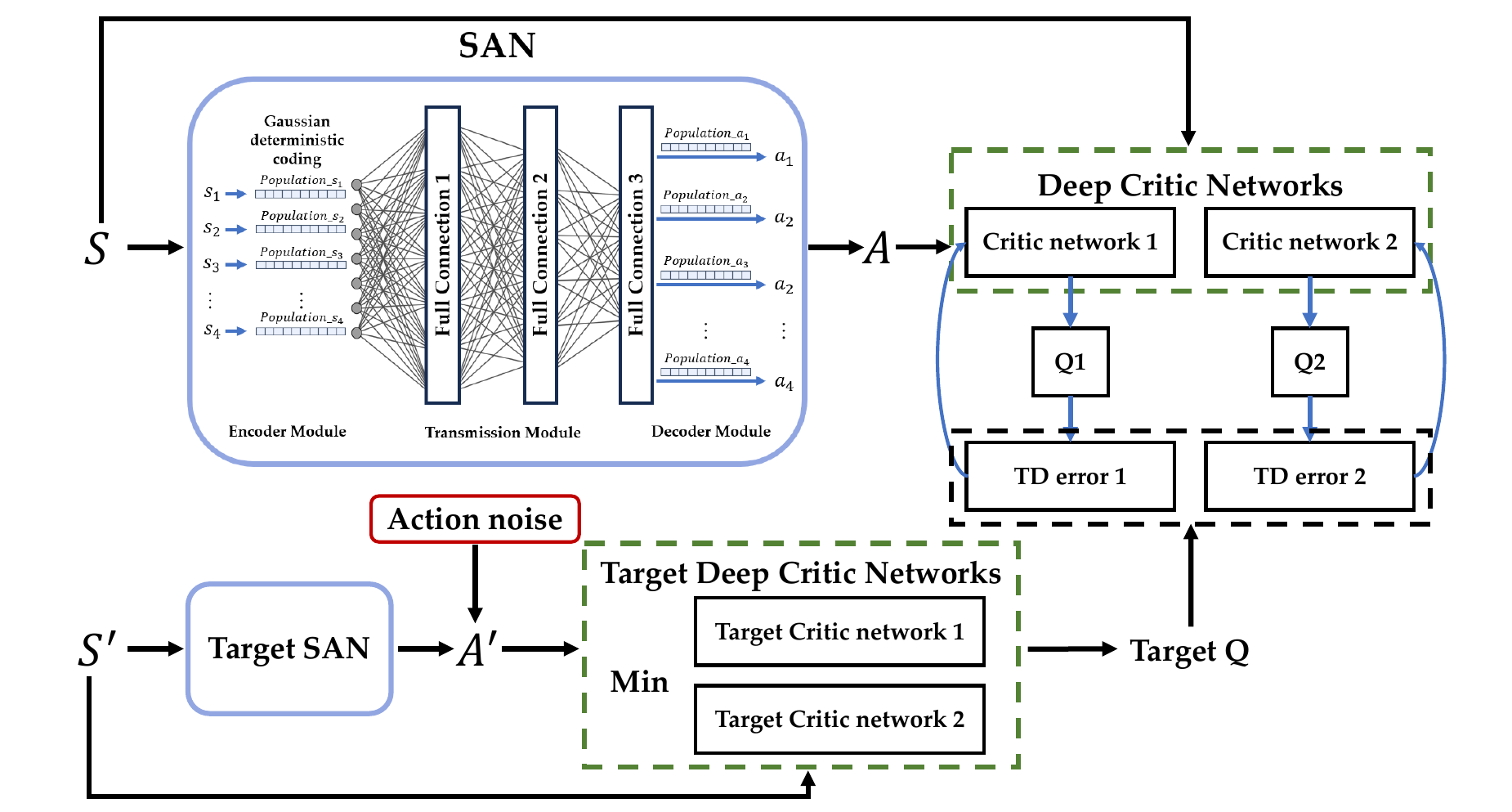}
   \vspace{-12pt}
  \caption {Introducing SAN algorithm framework into TD3 algorithm model based on AC framework, with the spiking actor network replacing the deep actor network.}
  \label{fig_system}
   \vspace{-18pt}
\end{figure*}



Based on the research of Pop-SAN and MDC-SAN, this paper proposes a new trapezoidal approximation gradient function. Considering that the simple functions used to replace the non-differentiable parts of spiking networks do not allow neuronal membrane potentials to sense the distance from the threshold during training, the new function addresses this issue. When the neuronal membrane potential is close to the threshold, the computed gradients are multiplied by a stable coefficient. When the membrane potential is far from the threshold, the gradient is multiplied by a coefficient that linearly decreases to zero with the distance from the threshold. This trapezoidal approximation gradient function not only improves the performance of the SAN algorithm but also enhances the stability of the model.



\section{SAN Algorithm Framework Overview}
\subsection{Introduction to the Reinforcement Learning Framework}
As illustrated in Fig.~\ref{fig_system}, both the Pop-SAN and MDC-SAN algorithms are modifications based on the AC framework. These algorithms replace the traditional deep actor network with a spiking actor network and train it in conjunction with a deep critic network, paving the way for the development of energy-efficient intelligent algorithms. To provide a foundation for the subsequent introduction of SAN algorithms \cite{fujimoto2018addressing}, we first introduce the classic Twin Delayed Deep Deterministic Policy Gradient (TD3) algorithm, which utilizes the AC framework.

The AC framework is a core architecture in RL, consisting of two parts: the actor network and the critic network. During the algorithm update process, a tuple $\langle s, a, r, s', d \rangle$ is sampled from the experience replay buffer. Here, $s$ denotes the state observed by the agent at time step $t$, $a$ represents the action taken based on state $s$, $r$ is the reward received after executing action $a$, $s'$ is the subsequent state observed at time step $t+1$, and $d$ indicates whether the task has done. The roles of actor network and the target actor network are to estimate the optimal actions $a$ and $a'$ based on the observed state information $s$ and $s'$, aiming to maximize Q-values. Noise is introduced to the action $a'$ during input to the target critic double network to enhance exploration. The critic double network outputs Q-values $Q_1$ and $Q_2$ based on $(s, a)$, while the target critic double network outputs $Q_1'$ and $Q_2'$ based on $(s', a')$. To address potential Q-value overestimation, TD3 employs the principle of Double Q-learning, selecting the smaller of the two estimated $Q'$ values, $Q_1'$ or $Q_2'$, as $Q_{\text{target}}$. This approach effectively limits the overestimation of Q-values by the Critic network, thereby ensuring the stability of the learning process.






The parameters of the critic double networks are updated through gradient descent backpropagation. Each network within the critic double networks updates independently. The TD error and the gradient descent equations can be expressed as follows:
\begin{equation}
   L(\theta) = \mathbb{E}\left[\left(r + \gamma \min_{i=1,2} Q'_{i}(s', \pi_{\phi}(s')) - Q_{\theta}(s, a)\right)^2\right]
\end{equation}
\begin{equation}
   \theta \leftarrow \theta - \alpha \nabla_{\theta} L(\theta)
\end{equation}
where $L(\theta)$ is the loss function, $\gamma$ is the discount factor, and $\theta$ represents the parameters of one of the networks in the critic, with the other network being updated similarly.





Subsequently, the parameters of the actor network are updated using a gradient ascent method to maximize the Q-values estimated by the critic network. The gradient ascent update equation for the parameters can be expressed as follows:
\begin{align}
\phi \leftarrow \phi + \alpha \nabla_\phi E\left[Q_\theta\left(s, \pi_\phi(s)\right)\right],
\end{align}
where $\alpha$ is the learning rate, $\nabla_\phi$ represents the gradient with respect to the parameter $\phi$, and $E\left[Q_\theta\left(s, \pi_\phi(s)\right)\right]$ is the expected Q-value, estimated by the critic network based on the current policy $\pi_\phi(s)$.

The target actor and critic networks employ a soft update mechanism to copy parameters from the actor and critic networks. The update equations can be expressed as follows:
\begin{align}
\theta' \leftarrow \tau\theta + (1 - \tau)\theta',  \\
\phi' \leftarrow \tau\phi + (1 - \tau)\phi',
\end{align}
where $\phi$ and $\phi'$ are the parameters of the actor and target actor networks, respectively, and $\theta$ and $\theta'$ are the parameters of the critic and target critic networks, respectively.


\subsection{SNN Playing the Role of the Actor Network}
In SAN algorithms, SNNs is used as the actor network to provide optimal actions based on input states, maximizing subsequent Q-values. 


\subsubsection{Encoder Module}
As shown in Fig.~\ref{fig_system}, the role of the encoder is to convert the input continuous state information into spike signals through population coding, meaning that the state information is represented as spike signals by multiple groups of neurons \cite{averbeck2006neural}. The total number of neurons in the encoder can be expressed as:
\begin{align}
 N_{enc} = N_o \times N_p,
\end{align}
where $N_o$ represents the dimension of the input state information, and $N_p$ represents the number of neuron groups.



The main tasks of the encoder are divided into two parts. The first part involves converting the state information into the stimulus intensity for each neuron. For the state information $s_i$, $i \in \{1, \ldots, N_o\}$, there is a group of neurons $E_{i,j}$ to receive it, $j \in \{1, \ldots, N_p\}$. The neurons in $E_{i,j}$ have Gaussian receptive fields $(\mu_{i,j}, \sigma_{i,j})$. The definition of the stimulus intensity $A_{E_{i,j}}$ can be expressed as:
\begin{align}
 A_{E_{i,j}} = e^{-\frac{1}{2}\left(\frac{s_i - \mu_{i,j}}{\sigma_{i,j}}\right)^2}.
\label{Gaussian}
\end{align}


The second part involves feeding the stimulus intensity $A_{E_{i,j}}$ into the specific neuron group, thereby inducing the neurons to generate spikes. The spiking activation mode of neuron groups can be expressed as follows:
\begin{align}
    v_{i,j}(t) &= v_{i,j}(t-1) + A_{E_{i,j}}, \label{eq:membrane_potential} \\
    o_{i,j}(t) &= 1, \quad \text{if } v_{i,j}(t) > 1 - \epsilon, \label{eq:spike_activation} \\
    v_{i,j}(t) &= v_{i,j}(t) - (1 - \epsilon), \quad \text{if } v_{i,j}(t) > 1 - \epsilon ,\label{eq:membrane_reset}
\end{align}
where $v_{i,j}(t)$ denotes the membrane potential of the $i$-th dimension's $j$-th neuron at time $t$, $1 - \varepsilon$ represents the encoder threshold, and $o_{i,j}(t)$ indicates the spike emission. Equation \eqref{eq:membrane_potential} updates each neuron's membrane potential based on \( A_{E_{i,j}} \). equations \eqref{eq:spike_activation} and \eqref{eq:membrane_reset} specify the conditions for spike generation: if a neuron's membrane potential exceeds the threshold $1 - \varepsilon$, a spike is emitted and the neuron's membrane potential resets.



\subsubsection{Transmission Module}
The information transmission module of the SNNs is implemented using a fully connected structure. The number of hidden layer neurons in this module differs from the total neuron count in both the encoder and decoder, necessitating dimension conversion through linear functions.

The transmission modules of the Pop-SAN and MDC-SAN algorithms employ different types of neurons: LIF neurons for Pop-SAN and more complex dynamic neurons for MDC-SAN. Despite these differences in neuron types, their core principle of information transmission remains similar. Taking the LIF neuron model as an example, the spike response time step of the SNNs transmission module is $T$ steps, and the spike neuron population information is transmitted across $K$ layers. The information transmission process can be expressed as follows:
\begin{align}
    c_{t}^{k} &= d_{c} \cdot c_{t-1}^{k} + W^{k} \cdot o_{t}^{k-1} + b^{k}, \\
    v_{t}^{k} &= d_{v} \cdot v_{t-1}^{k} \cdot (1 - o_{t-1}^{k}) + c_{t}^{k}, \\
    o_{t}^{k} &= \text{Threshold}(v_{t}^{k}),
\end{align}
The input current $c_{t}^{k}$ of each neuron consists of two parts: one part is the residual current $c_{t-1}^{k}$, which is the remaining portion of the membrane potential from the previous time step. The other part is the current caused by the spike $o_{t}^{k-1}$ output from the previous layer. Each neuron's state $(c, v, o)$ is influenced by its membrane potential $v_{t}^{k}$, updated based on the accumulated residual potential $v_{t-1}^{k}$ and the current input $c_{t}^{k}$. If the neuron spiked in the previous time step, it enters a refractory period. The spiking of a neuron depends on whether its membrane potential exceeds a predefined threshold. In the network, information is conveyed in the form of spikes.


\subsubsection{Decoder Module}
The role of the SNNs decoder module is to convert the activity patterns of the spiking neuron populations into comprehensible actions. In the design of the decoder module, the information of spiking neurons is decoded by accumulating the spike values from the output of the $k$-th layer of the SNNs over a time range $T$. The accumulation equation can be expressed as follows:
\begin{align}
    sc\ =\ \sum_{t=1}^{T}{\ o_t^K},
\end{align}
where $sc$ is a matrix with dimensions $N_a \times N_s$, recording the total response of neuron populations across all action dimensions $N_a$, with each neuron population having a size of $N_s$. Subsequently, the spike firing frequency is calculated for each action dimension's neuron population to represent its response. The spike firing frequency can be expressed as follows:
\begin{align}
    f^{\left(i\right)}\ =\ \frac{sc^{\left(i\right)}}{T}.
\end{align}

Subsequently, these accumulated values are transformed through a predefined weight matrix function that maps the discrete spike values to continuous action outputs. The weight matrix serves as a linear transformation layer, allowing the neural network to learn to adjust its action decisions based on the activity patterns of the spiking neural populations.The weight matrix can be expressed as follows:
\begin{align}
    a^i\ =\ W_a^{\left(i\right)} \cdot f^{\left(i\right)}\ +\ b_a^{\left(i\right)}.
    \label{16}
\end{align}


\section{Approximate Gradients To Replace The Spiking Network}
\subsection{Introduction to the Trapezoidal Approximate Gradient Function}
We utilize the approximate backpropagation algorithm to update the parameters in the SNNs via gradient descent. The core feature of approximate backpropagation is replacing the non-differentiable parts of the SNNs with a predefined approximate gradient function. In the MDC-SAN and Pop-SAN algorithms, a rectangular function is used to approximate the gradients of the spiking activity, achieving good training results. Common approximate gradient methods include \cite{wu2018spatio}: \textbf{Rectangular Gradient} which defines a constant gradient near the spiking threshold. It is simple but may cause training instability. \textbf{Triangular Gradient} whose gradient changes linearly near the threshold, forming a triangular gradient window. This provides a smooth gradient transition to enhance network learning stability.

\begin{figure}[ht]
  \centering
  \includegraphics[width=0.92\columnwidth]{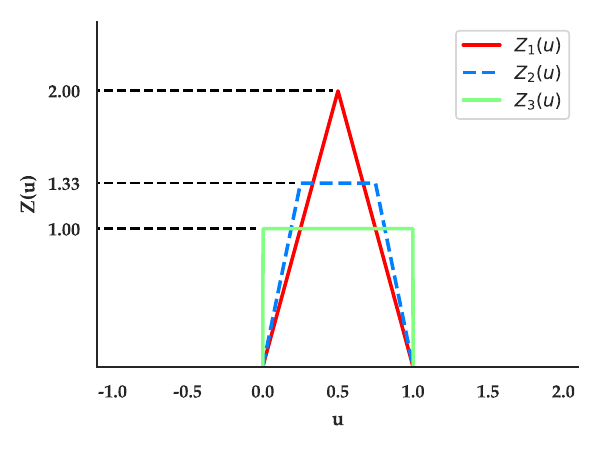}
   \vspace{-12pt}
  \centering \caption{Illustration of various approximate gradient functions: rectangular, triangular, and trapezoidal} 
  \label{trapezoidal}
   \vspace{-12pt}
\end{figure}

In this paper, as illustrated in Fig.~\ref{trapezoidal}, an innovative trapezoidal approximate gradient method is proposed based on an understanding of rectangular and triangular approximate gradients to improve the gradient approximation of spiking neurons. The trapezoidal function cleverly balances the stability of the rectangular function with the flexibility of the triangular function in its design. The rectangular function plays a key role in stabilizing model training due to its simplicity, while the triangular function has an advantage in model adaptability because of its flexibility. By combining the advantages of both functions, the trapezoidal function not only maintains stable learning outcomes but also enhances the model's adaptability and sensitivity to different signal dynamics. The trapezoidal approximate gradient function can be expressed as follows:
\begin{align}
z(V) = \begin{cases} h, & \text{if } |V - V_{\text{th}}| < w_1, \\h \left(1 - \frac{|V - V_{\text{th}}| - w_1}{w_2 - w_1}\right), & \text{if } w_1 \leq |V - V_{\text{th}}| < w_2, \\0, & \text{otherwise},\end{cases}
\label{obj2}
\end{align}
where $z$ is the trapezoidal approximate gradient function, $w_1$ is the width of the upper base of the trapezoid, and $w_2$ is the width of the lower base of the trapezoid. $V_{\mathrm{th}}$ is the discharge threshold, $V$ is the membrane voltage, and $h$ is the height of the trapezoid.  In the gradient approximation of spiking neurons, the trapezoidal function used should satisfy the condition that the overall integral is 1. This normalization condition ensures that the gradient approximation method will not introduce additional scaling factors during training.

\subsection{SNNs Parameters Update Process}
During the training of the SAN algorithm, the approximate gradient function replaces the non-differentiable part of the spiking network. The parameters of the spiking neuron layer can be updated based on the action loss gradient $\mathrm{\nabla}_{a_i}L$ calculated during backpropagation by the critic network. Next, I will sequentially introduce the parameter update methods for the decoder module, spiking neural network transmission module, and encoder module.

For the parameters update of the decoder module, as indicated by equation \eqref{16}, training is conducted using backpropagation gradient descent. The parameters $W_a^{(i)}$ and $b_a^{(i)}$ can be updated as follows:
\begin{align}
    \nabla_{w_a^{(i)}} L = \nabla_{a_i} L \cdot W_a^{(i)} \cdot f^{(i)}, \label{tidu -1} \\
    \nabla_{b_a^{(i)}} L = \nabla_{a_i} L \cdot W_a^{(i)}, \quad\quad\enspace\label{tidu -2}
\end{align}
equations \eqref{tidu -1} and \eqref{tidu -2} show how each group of neurons $i$ in the decoder module, $i \in \{1, \ldots, N_a\}$, is updated independently.

For the information propagation of the SNNs, SNNs parameters are updated through backpropagation through time (BPTT), using the trapezoidal function to approximate the gradients of the spiking network. By accumulating all temporal backpropagation gradients, the loss gradients with respect to SNNs parameters can be computed as follows:
\begin{align}
    \nabla_{w^k} L = \sum_{t = 1}^{T} o_t^{k-1} \cdot \nabla_{c_t^{k}} L, \label{tidu -3} \\
    \nabla_{b^k} L = \sum_{t = 1}^{T} \nabla_{c_t^{k}} L. \quad\quad\enspace \label{tidu -4}
\end{align}

For updating the encoder module parameters, backpropagation gradient descent \cite{lee2016training} can also be used as follows:
\begin{align}
    \nabla_{\mu_{i,j}} L &= \sum_{t = 1}^{T} \nabla_{{o_t^0}_{i,j}} L \cdot A_{E_{i,j}} \cdot \frac{s_i - \mu_{i,j}}{\sigma_{i,j}^2}, \label{tidu -5} \\
    \nabla_{\sigma_{i,j}} L &= \sum_{t = 1}^{T} \nabla_{{o_t^0}_{i,j}} L \cdot A_{E_{i,j}} \cdot  \frac{\left(s_i - \mu_{i,j}\right)^2}{\sigma_{i,j}^3} , \label{tidu -6}
\end{align}
equations \eqref{tidu -5} and \eqref{tidu -6} illustrate how each input neuron $i$ in the encoder, $i \in \{1, \ldots, N_0\}$, is updated independently.



\section{PERFORMANCE EVALUATION}
\subsection{Simulation Settings}

We first outline the configuration for our experiments within the SAN algorithm framework, where SNNs serve as replacements for the actor network in the AC framework. To ascertain the influence of various surrogate gradient methods on the performance of the algorithm, we maintained consistency across all hyperparameters to ensure experimental fairness. The learning rates for the actor and critic networks, denoted as $actor\_lr$ and $critic\_lr$, are set at 1e-4 and 1e-3 respectively, employing Adam as the optimizer. The neuron population sizes for the encoder ($encoder\_pop\_dim$) and decoder ($decoder\_pop\_dim$) are both set at 10, while the discount factor $\gamma$ is fixed at 0.99.

\subsection{Performance Comparison}
To thoroughly investigate the impact of different approximate gradient methods on algorithm performance, we conducted a comparative analysis between the trapezoidal approximate gradient method proposed in this paper and traditional rectangular and triangular approximate gradient methods. Comparative experiments are carried out in the HalfCheetah-v3 environment on the OpenAI Gym platform. Each approximate gradient method underwent 8 independent experiments to ensure the reliability and statistical significance of the results. All experiments were conducted under identical conditions to fairly reflect the impact of different gradient methods on SAN algorithm performance.

\begin{figure}[ht]
  \centering
  \includegraphics[width=0.92\columnwidth]{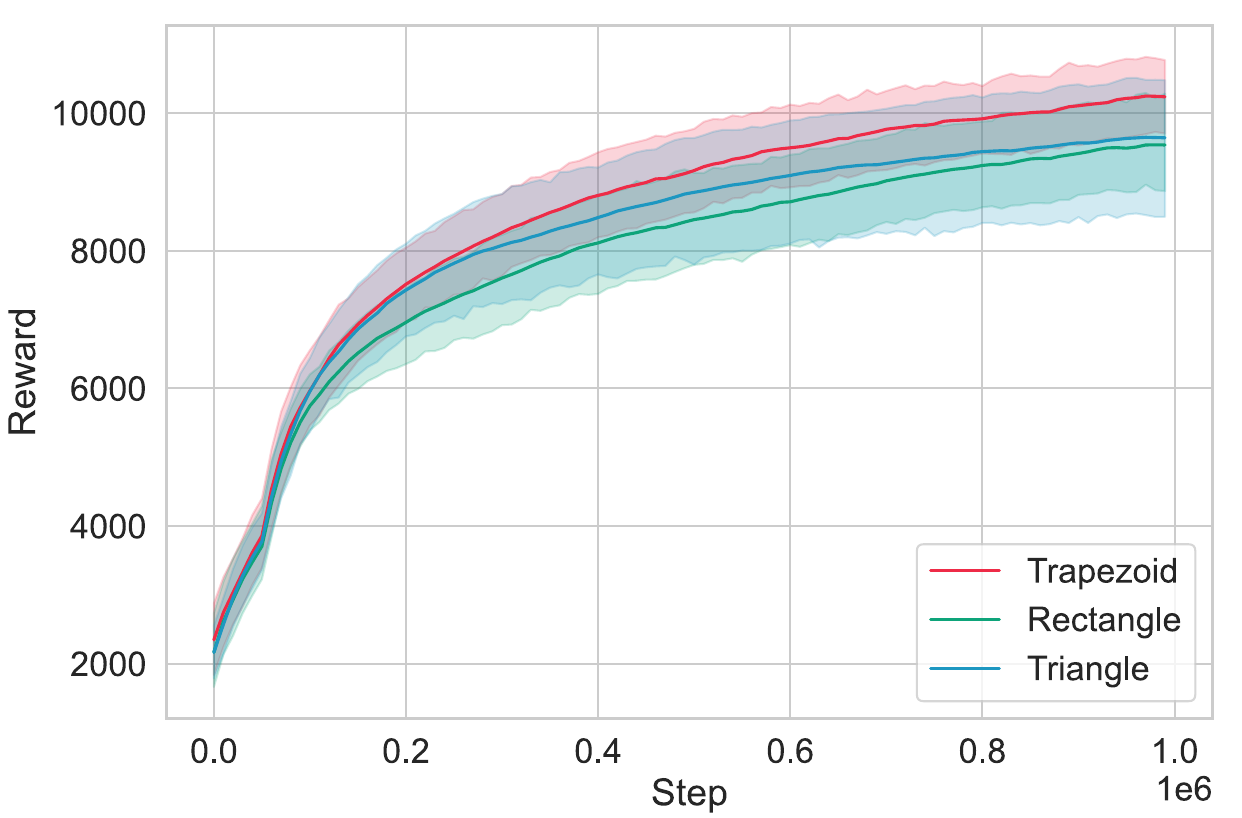}
   \vspace{-12pt}
  \centering \caption{The impact of different approximate gradient methods on Pop-SAN performance} 
  \label{Pop-SAN}
   \vspace{-12pt}
\end{figure}

As shown in Fig.~\ref{Pop-SAN}, during the early stages of training, the convergence speed of the triangular approximate gradient is similar to that of the trapezoidal approximate gradient. However, as training progresses, the performance of the triangular approximate gradient gradually aligns with that of the rectangular approximate gradient but remains inferior to the trapezoidal approximate gradient we proposed. Meanwhile, from the Fig.~\ref{Pop-SAN}, it can be observed that the variance of the trapezoidal gradient curve's shaded region is much smaller than that of the other two methods, indicating higher stability in algorithm training.


\begin{figure}[ht]
  \centering
  \includegraphics[width=0.92\columnwidth]{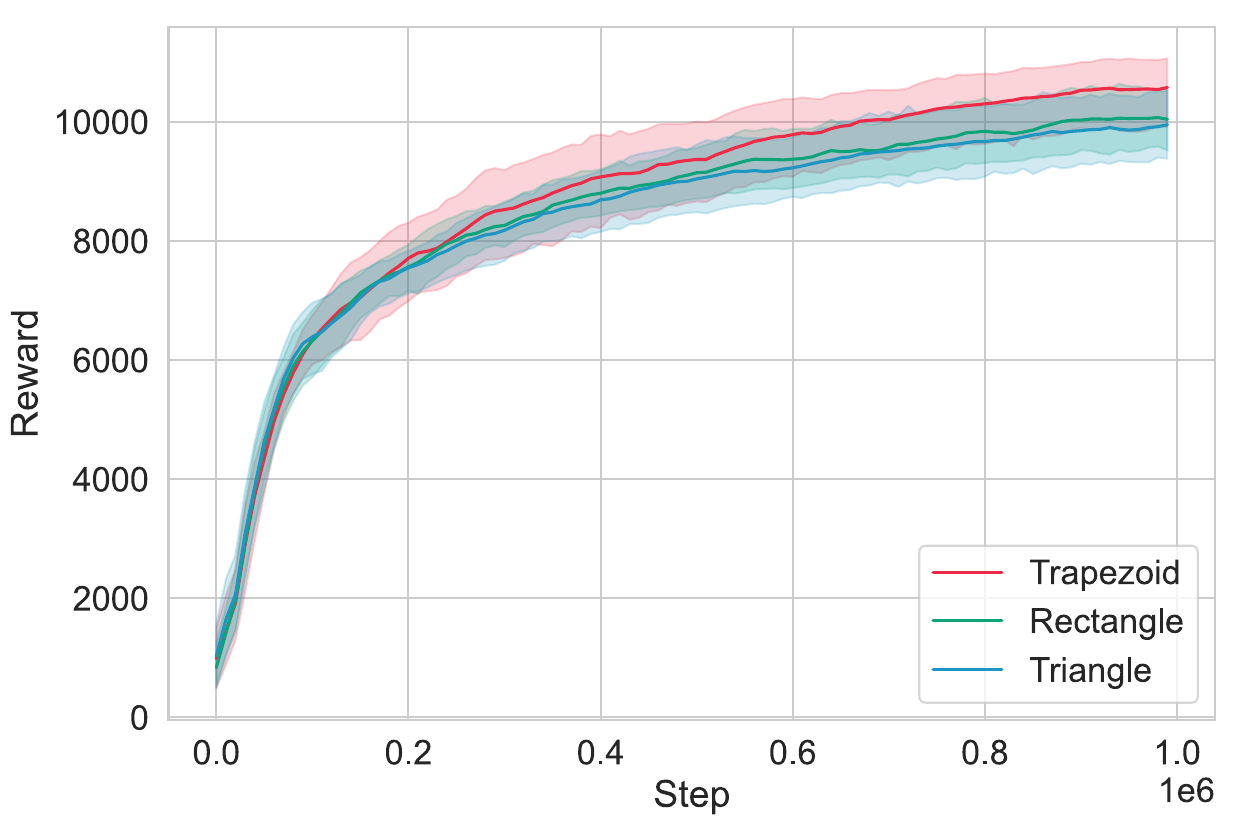}
   \vspace{-12pt}
  \centering \caption{The impact of different approximate gradient methods on MDC-SAN performance} 
  \label{MDC-SAN}
   \vspace{-15pt}
\end{figure}

As shown in Fig.~\ref{MDC-SAN}, the reward curves of the three methods almost overlap in the early training stages. As training progresses, the trapezoidal approximate gradient takes the lead, surpassing the other methods in reward, demonstrating its effectiveness in policy convergence and performance improvement. In the later training stages, it also maintains excellent stability and consistency.


Therefore, selecting an appropriate approximate gradient function to replace the spiking network can enhance the performance of spiking reinforcement learning algorithms and improve the training effectiveness of SNNs. This enhancement is due to the trapezoidal approximate gradient function's better simulation of neuronal electrophysiology. By providing a smooth gradient response near the neuron threshold and a gradually diminishing response further away, the model's adaptability and sensitivity to signal changes are increased.


\section{Conclusion}
In this paper, we introduce a new trapezoidal approximation gradient function to replace the non-differentiable part of spiking neurons, aiming to address the insensitivity issue of spiking networks near the threshold. Compared to traditional rectangular and triangular functions, this new gradient approximation method shows notable advantages in convergence performance. In future research, we will explore innovations in the structure of SNNs, such as investigating more efficient ways for neurons to connect.

\bibliography{ref}
\bibliographystyle{IEEEtran}

\ifCLASSOPTIONcaptionsoff
  \newpage
\fi

\end{document}